\pdfoutput=1
\documentclass[sigconf,dvipsnames]{acmart}
\usepackage{tabularx}
\usepackage{booktabs}
\usepackage{algorithm}
\usepackage{algorithmicx}
\usepackage{algpseudocode}  
\usepackage{colortbl}
\usepackage{epsfig}
\usepackage{graphicx}
\usepackage{amsmath}
\usepackage{amssymb}
\usepackage{enumitem}
\usepackage{xcolor}
\usepackage{xspace}
\usepackage{balance}
\usepackage{multirow}
\usepackage{mathtools}
\usepackage{rotating}
\usepackage{url}
\usepackage{tabulary}
\usepackage{array}
\usepackage[export]{adjustbox}
\usepackage{ctable}
\usepackage{subcaption}

\def\etc{\emph{etc.}}
\def\Vec#1{{\boldsymbol{#1}}}
\def\Mat#1{{\boldsymbol{#1}}}

\definecolor{myblue1}{RGB}{0,127,255}
\def\CellB#1{\cellcolor{myblue1!25}#1}

\newcommand{\fig}{{Figure}\@\xspace}
\newcommand{\tab}{{Table}\@\xspace}
\newcommand{\eqn}{{Eq.}\@\xspace}

\newcommand{\ie}{{i.e.}\@\xspace}
\newcommand{\eg}{{e.g.}\@\xspace}
\newcommand{\etal}{{\it et~al.}\@\xspace}
\DeclareMathOperator*{\argmin}{\arg\min}
\DeclareMathOperator*{\argmax}{\arg\max}

\graphicspath{{./figs/}}

\begin{document}

\fancyhead{}
\settopmatter{printacmref=false, printfolios=false}

\title
	{
	Attention Transfer from Web Images for Video Recognition
	}

	\author{Junnan~Li}
	\affiliation{%
		\institution{NUS Graduate School for Integrative Sciences and Engineering, National University of Singapore}
	}
	\email{lijunnan@u.nus.edu}
	
	\author{Yongkang~Wong}
  \orcid{0000-0002-1239-4428}
	\affiliation{%
		\institution{Interactive \& Digital Media Institute\\National University of Singapore}
	}
	\email{yongkang.wong@nus.edu.sg}

	\author{Qi~Zhao}
	\affiliation{%
		\institution{Department of Computer Science and Engineering\\University of Minnesota}
	}
	\email{qzhao@cs.umn.edu}

	\author{Mohan~S.~Kankanhalli}
	\affiliation{%
		\institution{School of Computing\\National University of Singapore}
	}
	\email{mohan@comp.nus.edu.sg}

\begin{abstract}
	
Training deep learning based video classifiers for action recognition requires a large amount of labeled videos.
The labeling process is labor-intensive and time-consuming.
On the other hand,
large amount of weakly-labeled images are uploaded to the Internet by users everyday.
To harness the rich and highly diverse set of Web images,
% a scalable approach is to use commercial search engines to crawl images for classifier training.
a scalable approach is to crawl these images to train deep learning based classifier,
such as Convolutional Neural Networks (CNN).
% With the development of commercial search engines,
% crawling Web images to train deep classifiers is a scalable approach.
However, 
due to the domain shift problem,
the performance of Web images trained deep classifiers tend to degrade when directly deployed to videos.
One way to address this problem is to fine-tune the trained models on videos, 
but sufficient amount of annotated videos are still required.
In this work, 
we propose a novel approach to transfer knowledge from image domain to video domain.
The proposed method can adapt to the target domain (\ie~video data) with limited amount of training data.
Our method maps the video frames into a low-dimensional feature space using the class-discriminative spatial attention map for CNNs.
We design a novel Siamese EnergyNet structure to learn energy functions on the attention maps by jointly optimizing two loss functions,
such that the attention map corresponding to a ground truth concept would have higher energy.
We conduct extensive experiments on two challenging video recognition datasets (\ie~TVHI and UCF101), and demonstrate the efficacy of our proposed method.
% We apply our method to action/interaction recognition, and show that it achieves performance improvement on two challenging datasets.

\end{abstract}

\copyrightyear{2017} 
\acmYear{2017} 
\setcopyright{acmcopyright}
\acmConference{MM '17}{October 23--27, 2017}{Mountain View, CA, USA}\acmPrice{15.00}\acmDOI{10.1145/3123266.3123432}
\acmISBN{978-1-4503-4906-2/17/10}



% 	<ccs2012>
% 	<concept>
% 	<concept_id>10010147.10010178.10010224</concept_id>
% 	<concept_desc>Computing methodologies~Computer vision</concept_desc>
% 	<concept_significance>500</concept_significance>
% 	</concept>
% 	<concept>
% 	<concept_id>10010147.10010178.10010224.10010225.10010228</concept_id>
% 	<concept_desc>Computing methodologies~Activity recognition and understanding</concept_desc>
% 	<concept_significance>300</concept_significance>
% 	</concept>
% 	<concept>
% 	<concept_id>10010147.10010257.10010293.10010294</concept_id>
% 	<concept_desc>Computing methodologies~Neural networks</concept_desc>
% 	<concept_significance>300</concept_significance>
% 	</concept>
% 	</ccs2012>
% \ccsdesc[500]{Computing methodologies~Computer vision}
% \ccsdesc[300]{Computing methodologies~Activity recognition and understanding}
% \ccsdesc[300]{Computing methodologies~Neural networks}


\maketitle

\section{Introduction}
\label{sec:introduction}

Recent advancements in deep Convolutional Neural Network (CNN) have led to promising results in large-scale video classification~\cite{Donahue_CVPR_2015,Karpathy_CVPR_2014,Simonyan_NIPS_2014,Tran_ICCV_2015,Wang_CVPR_2015}. 
A prerequisite of deep model training is the availability of large-scale labeled training data. 
% high quality annotated training videos. 
However, 
the acquisition and annotation of such datasets (e.g. UCF101~\cite{Soomro_CORR_2012}, ActivityNet~\cite{Heilbron_CVPR_2015}, Sport1M~\cite{Karpathy_CVPR_2014}) is often labor-intensive.
On the other hand, Web images are easier to collect by querying widely available commercial search engines. 
Unlike videos,
%  that contain redundant information, 
Web images usually capture the representative moments of events or actions and provide more diverse examples for each concept.
This makes them a good auxiliary source to enhance video concept recognition.

\begin{figure}[!t]
 \centering
  \begin{minipage}{1.0\columnwidth}  	
%   	\resizebox{\columnwidth}{!}{%
%   		\includegraphics[height=3.5cm]{handshake_web}
%   		\includegraphics[height=3.5cm]{handshake_video}
%   	}
%   	\vspace{-4ex}
%   	\subcaption{\footnotesize Handshake}
%   	\vspace{1ex}
  	\resizebox{\columnwidth}{!}{%
  		\includegraphics[height=3.5cm]{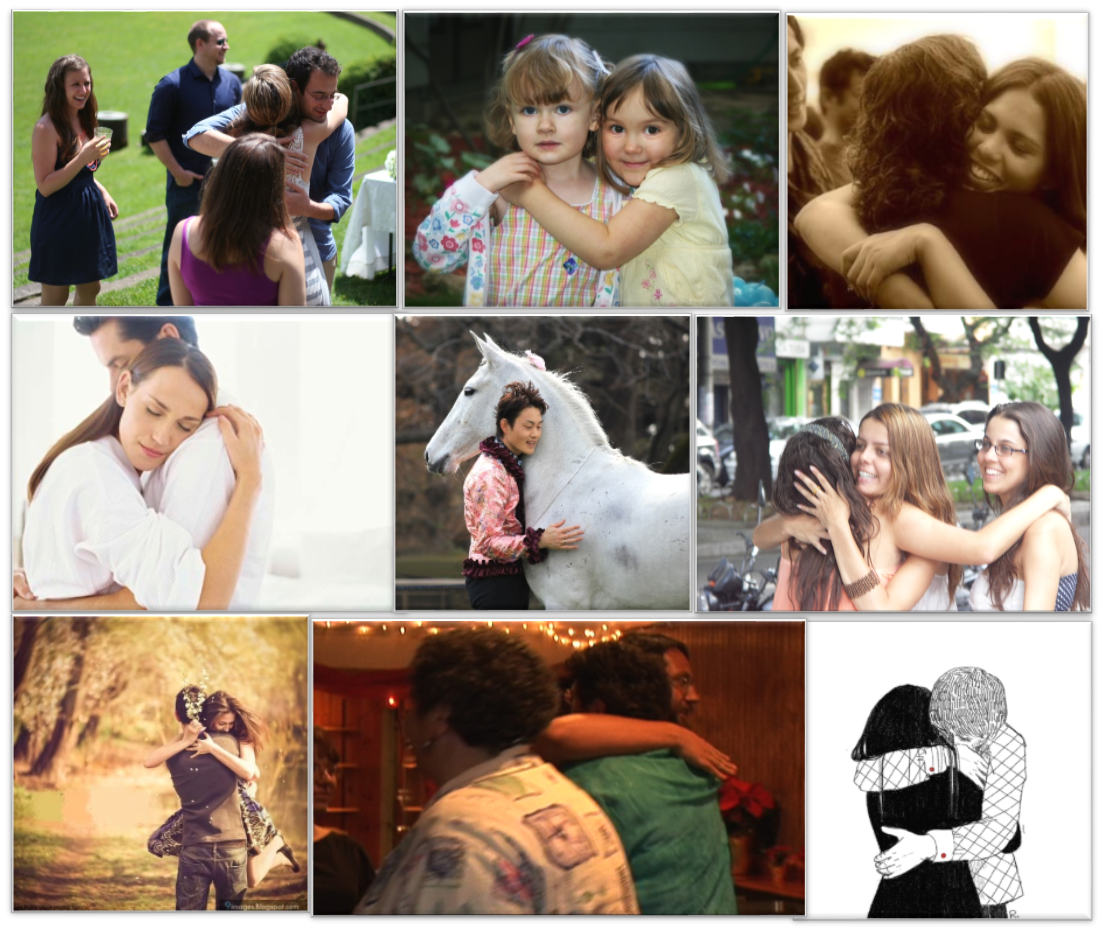}
  		\includegraphics[height=3.5cm]{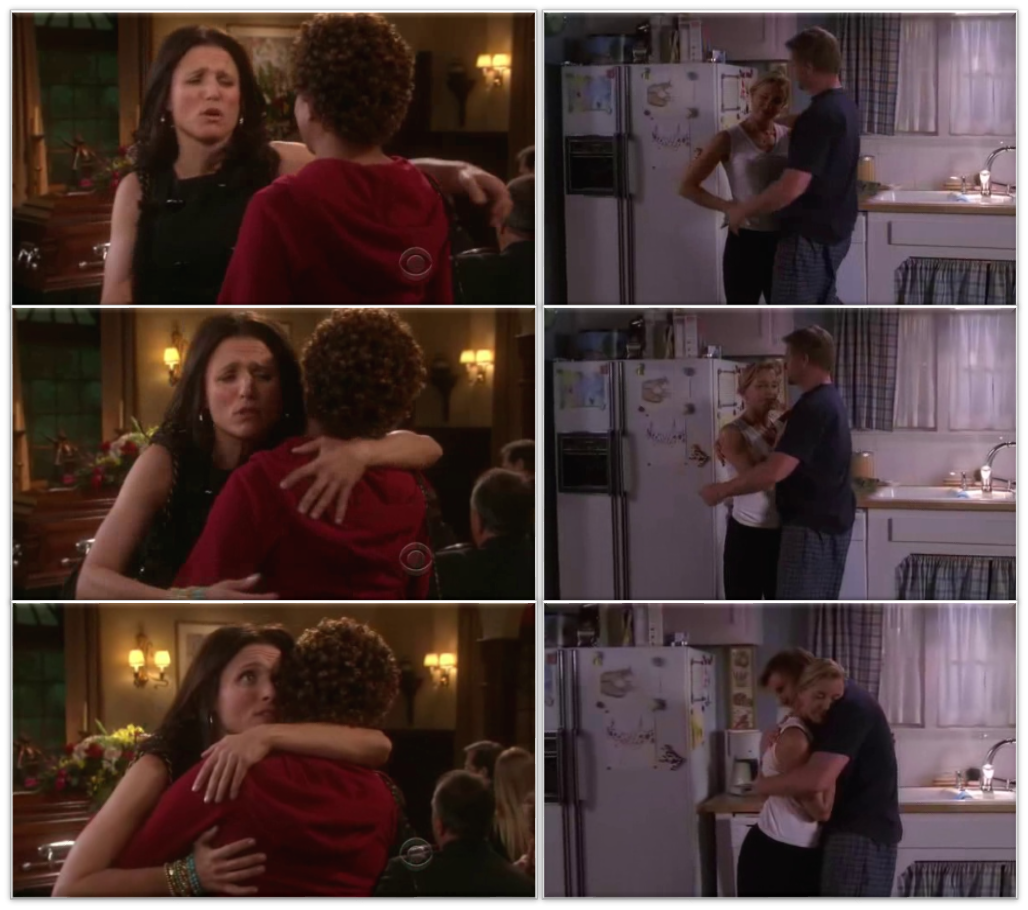}
  	}
%   	\vspace{-4ex}
%   	\subcaption{\footnotesize Hug}
%  	\vspace{1ex}
%  	\resizebox{\columnwidth}{!}{%
%  		\includegraphics[height=3.5cm]{kiss_web}
%  		\includegraphics[height=3.5cm]{kiss_video}
%  	}
%  	\vspace{-4ex}  	
%  	\subcaption{\footnotesize Kiss}
  \end{minipage}
  \vspace{-2ex}
  \caption
    {
   	\small
	Weakly-labeled Web images collected from commercial search engines provide rich and diverse training data for model training. 
	However, 
	there exist domain shift between Web images and videos, 
	where the data distributions of the two domains differ. 
	(Left: Web images retrieved by keywords. Right: Video frames from
	TVHI dataset.) 
	}
  \label{fig:samples}
\end{figure}

A naive approach to harness information from Web images is to directly apply the Web images trained classifier to video data.
However, 
learning video concepts from Web images introduces the domain shift problem~\cite{Saenko_ECCV_2010},
where the variation in data between the source and target domain jeopardize the performance of the trained classifier.
%  the classifier learned from the source domain does not perform well on the target domain because of different data distributions between the two domains.
As shown in \fig~\ref{fig:samples}, the Web images (left) differ from video frames (right) in background, color, lighting, actors, and so on.
Several recent works~\cite{Gan_CVPR_2016,Ma_PR_2017,Sun_MM_2016,Gan_ECCV_2016} address this by jointly utilizing images and videos to train shared CNNs, 
and use the shared CNNs to map both images and videos into the same feature space. 
% represented by the features of the last layer of the CNN.
However, 
in order to learn domain-invariant feature representations for the shared CNNs,
sufficient amount of annotated video data are required for training.
It is expensive and time-consuming to collect labeled training data for various video domains (\eg movies, consumer videos, egocentric videos, \etc).

\begin{figure*}[!t]
 \centering
  \begin{minipage}{1.0\textwidth}
  	\begin{minipage}{0.195\columnwidth}
		 	\centerline{\includegraphics[width=\linewidth]{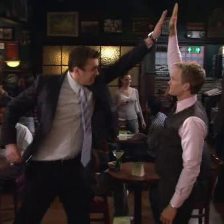}}
		 	\subcaption{\small Original Frame}
  	\end{minipage}
  	\hfill
  	\begin{minipage}{0.195\columnwidth}
  		\centerline{\includegraphics[width=\linewidth]{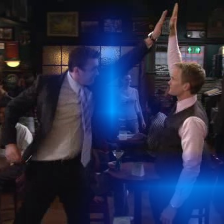}}
  		\subcaption{\small Handshake}
  	\end{minipage}
  	\hfill
  	\begin{minipage}{0.195\columnwidth}
  		\centerline{\includegraphics[width=\linewidth]{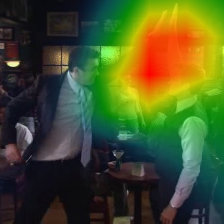}}
  		\subcaption{\small High Five}
  	\end{minipage}
  	\hfill
  	\begin{minipage}{0.195\columnwidth}
  		\centerline{\includegraphics[width=\linewidth]{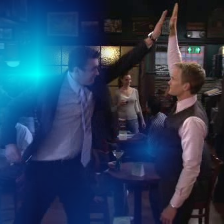}}
  		\subcaption{\small Hug}
  	\end{minipage}
  	\hfill
  	\begin{minipage}{0.195\columnwidth}
  		\centerline{\includegraphics[width=\linewidth]{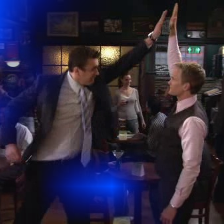}}
  	    \subcaption{\small Kiss}
  	\end{minipage}	
  \end{minipage}
  \vspace{-2ex}
  \caption
    {
		\small
		Examples of heatmap overlayed on a video frame.
		The respective spatial attention maps are generated with a Web image trained CNN.
    } 
    \vspace{-2ex}
  \label{fig:attmap}
\end{figure*}

There exist two common scenarios to adapt the classifier trained on a source domain to the target domain:
(1)~{\it unsupervised} scenario: no labeled data in the target domain is available; and 
(2)~{\it supervised} scenario: labeled training data is available in the target domain.
In this work, we propose a new approach to transfer knowledge from Web images to videos.
Comparing with using the features from the last layers of CNNs for video recognition~\cite{Gan_CVPR_2016,Ma_PR_2017,Sun_MM_2016,Gan_ECCV_2016}, 
our method can achieve better performance on target domain under both unsupervised and supervised scenarios.
%Given training data in the target domain, our method can also better address the domain shift problem.
%Given limited amount of training data in the target domain, our method can also better address the domain shift problem.

The proposed approach is based upon class-discriminative spatial attention maps for CNNs~\cite{Selvaraju_CORR_2016,Zagoruyko_ICLR_2017,Zhou_CVPR_2016},
which are initially proposed to visualize the discriminative regions on images that are `important' for each class prediction.
The spatial attention maps are later applied for weakly-supervised object localization~\cite{Selvaraju_CORR_2016,Zhou_CVPR_2016},
and to transfer knowledge from deeper networks to shallower networks~\cite{Zagoruyko_ICLR_2017}.

Our method roots from this observation: 
Given a video frame and a Web images trained concept classifier (\ie~CNN model),
if a concept appears in that frame, 
certain regions in the spatial attention map w.r.t. that concept would have high energy heatmap (or activation).
On the other hand, if a concept does not appear in the frame,
the corresponding spatial attention map would have low and sparse activations (see \fig~\ref{fig:attmap}).
Therefore, the spatial attention maps are informative of the concept.
Furthermore, since the spatial attention map is computed from the features of the convolutional layer of CNNs (see Section~\ref*{sec:gradcam} for details),
we presume that it is more domain-invariant compared with the features of the last fully-connected (fc) layer.
In this study, we propose approaches to exploit the spatial attention map,
so that the video classifier trained on Web images would suffer less from the domain shift.

Our contributions are as follows:
\begin{itemize}
	\item 
	We propose to use class-discriminative spatial attention maps for cross-domain knowledge transfer from Web images to videos. 
	Experiments on action/interaction recognition with two challenging datasets (\ie~TVHI and UCF101) demonstrate the efficacy of the proposed methods.
	\item 
	We propose an energy-based method on the spatial attention maps, with the aim of assigning the highest energy to the ground truth concept.
	We design an Energy Network to learn class-specific energy functions, and construct a Siamese structure that jointly optimizes over two loss functions, energy loss and triplet loss.
	We show that our method can achieve superior performance over several baselines.
% 	 including a directly unsupervised approach.
	\item
	We collected a new Human Interaction Image (HII) dataset to facilitate research in interaction recognition, which contains images for four types of interactions (please refer to Section \ref{sec:dataset} for more details).  
\end{itemize}	

The rest of the paper is organized as follows:
Section~\ref{sec:literature} reviews the related work.
Section~\ref{sec:method} delineates the details of the proposed method.
Section~\ref{sec:experiments} elaborates on the experiments and discusses the results. 
Section~\ref{sec:conclusion} concludes the paper.
\section{Related Work}
\label{sec:literature}

\subsection{Action Recognition on Unconstrained Data}

Action recognition is a very active research area and has been widely studied.
A detailed survey can be found in~\cite{Cheng_CORR_2015}. 
Most existing works take a two-step approach: feature-extraction and classifier training.
Many hand-crafted features are designed for video appearance and motion representations,
where Improved Dense Trajectories (IDT)~\cite{Wang_ICCV_2013} combined with Fisher vector coding~\cite{Oneata_ICCV_2013} achieve state-of-the-art performance.
Recent approaches use deep networks (particularly CNNs) to jointly learn feature extractors and classifiers for action recognition. 
Tran~\etal~\cite{Tran_ICCV_2015} use 3D CNNs to learn spatial-temporal features.
Simonyan and Zisserman~\cite{Simonyan_NIPS_2014} propose two-stream networks:
one stream captures spatial information from video frames and the other stream captures motion information from stacked optical flows.
Recurrent Neural Networks (RNNs), with the ability to model sequential information,
have also been utilized for action recognition~\cite{Donahue_CVPR_2015,Ng_CVPR_2015}.
However, these deep learning based approaches all require large amount of well-labeled videos to avoid overfitting.

\subsection{Learning from Web Data}

To harness the information from large-scale Web images,
several works use Web images as auxiliary training data for video recognition~\cite{Ma_PR_2017,Sun_MM_2016,Gan_CVPR_2016,Gan_ECCV_2016}.
Ma~\etal~\cite{Ma_PR_2017} collect a large web action image dataset, 
and achieve performance gain by combining web images with video frames to train CNNs.
Sun~\etal~\cite{Sun_MM_2016} propose a domain transfer approach for action localization,
where they iteratively train a shared CNN on video frames and Web images.
Gan~\etal~\cite{Gan_CVPR_2016,Gan_ECCV_2016} jointly exploit Web images and Web videos, 
and propose a mutually voting approach to filter noisy Web images and video frames~\cite{Gan_ECCV_2016}.
%In~\cite{Gan_CVPR_2016}, they use a iterative training approach to filter out noisy images and irrelavant frames, and deploy a Long Short-Term Memory (LSTM) network to incorporate temporal information.
%In~\cite{Gan_ECCV_2016}, they propose a mutually voting approach to filter Web images and video frames.

Another common usage for Web images is to learn semantic concept detectors and apply them for video retrieval~\cite{Chen_ICMR_2014,Wu_CVPR_2014,Singh_CVPR_2015,Divvala_CVPR_2014}.
Chen~\etal~\cite{Chen_ICMR_2014} discover concepts from tags of Web images,
whereas Singh~\etal~\cite{Singh_CVPR_2015} construct pairs of concepts to crawl web images for training concept detectors.
However, 
due to the domain shift between Web images and videos, 
their detectors are not suitable for zero-shot video recognition and required to be retrained on videos.

Web images are inherently noisy. 
Several solutions are proposed to train CNNs on noisily labeled data~\cite{Sukhbaatar_CoRR_2014,Xiao_CVPR_2015,Chen_ICCV_2015}.
However, 
recent studies show that state-of-the-art CNNs trained with large-scale noisily labeled images are surprisingly effective in a range of vision problems~\cite{Joulin_ECCV_2016,Krause_ECCV_2016}.
This suggests that learning from weakly-labeled Web images is a scalable solution to train deep networks.

\subsection{Domain Adaptation}

Domain shift refers to the situation where data distribution differs between source domain and target domain, causing the classifier learned from source domain to perform poorly on target domain.
A large number of domain adaptation approaches have been proposed to address this problem,
where the key focus is to learn domain-invariant feature representations.
There are two common strategies:
one approach is to reweight the instances from the source domain~\cite{Chen_NIPS_2011,Long_CVPR_2014},
and the other approach is to find a mapping function that would align the source distribution with the target domain~\cite{Gong_CVPR_2012,Baktashmotlagh_ICCV_2013,Ni_CVPR_2013,Fernando_ICCV_2013}.

Deep networks can learn nonlinear feature representations that manifest the underlying invariant factors and are transferable across domains and tasks~\cite{Yosinski_NIPS_2014}.
Therefore, deep networks have been recently exploited for domain adaptation.
Tzeng~\etal~\cite{Tzeng_ICCV_2015} introduce an adaptation layer and domain confusion loss to learn domain-invariant features across tasks.
Bousmalis~\etal~\cite{Bousmalis_NIPS_2016} propose domain separation networks that explicitly model the unique characteristics for each domain,
so that the invariance of the shared feature representation is improved.
Long~\etal~\cite{Long_ICML_2015} build a deep adaptation network (DAN) that explores multiple kernel variant of maximum mean discrepancies (MK-MMD) to learn transferable features.
Yosinski~\etal~\cite{Yosinski_NIPS_2014} explore feature transferability of deep CNNs.
They show that while the first layers of a CNN can learn general features, the features from the last layers are more specific and less transferable.
Therefore the CNN needs to be fine-tuned on sufficient labeled target data to achieve domain adaptation.
Very recently, 
attention map has been studied as a mechanism to transfer knowledge~\cite{Zagoruyko_ICLR_2017}.
Different from the above problem, 
their work studies knowledge transfer from a deeper network to a shallower network within the same domain.

In this work, 
we explore the use of attention for cross-domain knowledge transfer from Web images to videos. 
We show that attention is a more transferable feature compared with the features from the last layers of CNN. 
Different from previous works that utilize Web images for video recognition~\cite{Ma_PR_2017,Sun_MM_2016,Gan_CVPR_2016,Gan_ECCV_2016}, 
our method can better address the domain shift problem with significantly less training data in the target domain.
 
\section{Proposed Method}
\label{sec:method}

In this section, 
we first briefly overview the Gradient-weighted Class Activation Mapping (Grad-CAM)~\cite{Selvaraju_CORR_2016},
which is the fundamental component that allows effective domain adaptation from Web image to video.
Then, 
we state the problem statement and detail the proposed domain adaptation approaches.

\subsection{Spatial Attention Map}
\label{sec:gradcam}

In this work,
we adopt Grad-CAM to generate class-discriminative spatial attention maps. 
Grad-CAM improves upon CAM~\cite{Zhou_CVPR_2016} and required no re-training of the CNN.
It has been shown to be a robust method for visualizing deep CNNs, and achieves state-of-the art results for weakly-supervised concept localization in images. 

Grad-CAM works as following.
Assume that we have a probe image (or image frame from a video sequence) $\Mat{F}$, a target concept $c$, and a trained CNN model,
which the last convolutional layer produce $K$ feature maps {\small $\Mat{A}^k$}.
The image $\Mat{F}$ is first forwardly propagated through the trained CNN model,
then Grad-CAM generates the spatial attention map {\small $L(\Mat{F},c)$} by a weighted combination of the convolutional feature maps,
\begin{equation}
	L(\Mat{F},c) = ReLU \big( \sum\nolimits_{k} \alpha_{k}^{c} \Mat{A}^{k} \big).
\end{equation}

The weight {\small $\alpha_{k}^{c}$} captures the importance of the $k$-th feature map for the concept $c$,
and is calculated by backpropagating gradients to the convolutional feature map {\small $\Mat{A}^{k}$}.
Prior the backpropagation operation, 
vector quantization is performed on the gradients for the penultimate layer of the CNN model 
(the layer before softmax that outputs raw scores)
where the dimension of concept $c$ is set to 1 and the remaining as 0.
The gradients flowing back to {\small $\Mat{A}^{k}$} are global-average-pooled to obtain {\small $\alpha_{k}^{c}$}.
More details of Grad-Cam can be found in \cite{Selvaraju_CORR_2016}.

%In this paper, 
%we adapt two widely-used CNN models, 
%namely ResNet~\cite{He_CVPR_2016} and VGGNet~\cite{Simonyan_CORR_2014},
%where the size of the attention map is fixed at $7\times7$ and $14\times14$, 
%respectively.
%\NOTE{move to other section?}
%\NOTE{I agree here is not so good, but where?}
%
%
  
\subsection{Problem Statement}

Given a set of weakly-labeled Web images and a set of videos that share the same set of concepts {\small $C=\{c_1,c_2,...,c_n\}$},
we first adopt state-of-the-art CNN models 
%(\ie~VGGNet~\cite{Simonyan_CORR_2014} and ResNet~\cite{He_CVPR_2016})
to pre-train the image-based classifier on the Web images.
In this work,
instead of using the given videos to train a video-based classifier,   
our goal is to study how to exploit the Web images trained CNN model to classify the videos.
Specifically,
we aim to explore domain adaptation mechanism so that the pre-trained model can better adapt to the domain shift when being deployed to videos.
We propose to address this problem by using the spatial attention maps {\small $L(\Mat{F},c_i)$} where {\small $i=1,2,\ldots,n$} under two scenarios.
Briefly,
the first scenario,
namely {\it unsupervised domain adaptation}, 
directly uses the Web images trained CNN model to the video frames without further training on any videos (Section~\ref{sec:uda}),
whereas the second scenario,
namely {\it supervised domain adaptation}, 
utilizes the available training videos to improve the domain adaptation (Section~\ref{sec:sda}).

\subsection{Unsupervised Domain Adaptation}
\label{sec:uda}
Directly applying the Web images trained CNN to classify videos would lead to poor performance.
This is because the score generated from the last fc layer of a CNN is domain-specific~\cite{Yosinski_NIPS_2014}.
Therefore, we propose to exploit features from the convolutional layer, using spatial attention map.
The attention map incorporates the more general convolutional feature with class information, hence is more transferable across domains.

For a frame $\Mat{F}$ in a given video, 
let {\small $L(\Mat{F},c_i)$} be the spatial attention map generated by a pre-trained CNN model for concept $c_i$.
Denote {\small $c_{\texttt{gt}}^\Mat{F}$} as the ground truth concept that exists in $\Mat{F}$,
we define an energy function $E$ on the spatial attention map, 
such that for each video, {\small $\sum_F{E(L(\Mat{F},c_i))}$} is the largest when {\small $c_i = c_{\texttt{gt}}^\Mat{F}$}, and smaller otherwise. 

We define $E$ based on a simple yet effective observation: 
Assuming that the CNN model has been pre-trained on Web images to detect certain concepts,
given a frame and its spatial attention map corresponding to a concept,
certain region of the frame would have higher activations in the attention map if the concept is present in the corresponding region (See \fig~\ref{fig:attmap}).
Therefore, 
we apply a sliding window over {\small $L(\Mat{F},c_i)$} with window size of $s \times s$ ($s=3$) and step size of 1.
We then compute the sum of the value of {\small $L(\Mat{F},c_i)$} within each sliding window as the local activation.
The maximum of all local activations is taken as the energy $E$.
For a video with $N$ frames, the output score over each concept is calculated as the mean energy across all frames,
\begin{equation}
	score(\Mat{F},c_i) = \dfrac{1}{N}\sum\nolimits_F{E(L(\Mat{F},c_i))},
\label{eqn:unsup_score}
\end{equation}

\noindent
and the predicted video-level concept is inferred as the one with the highest score,
\begin{equation}	
	c_{v} = \argmax_{c_i} \; score(\Mat{F},c_i).
\end{equation}

\subsection{Supervised Domain Adaptation}
\label{sec:sda}

Since each concept may have different activation patterns in the corresponding attention maps, 
a universal energy function proposed in Section \ref{sec:uda} could not optimally fit to all concepts.
Given training data from the video domain, 
we design an Energy Network (EnergyNet) to learn an energy function {\small $E_\texttt{net}(\Mat{F},c)$} that explicitly encodes class information.
In other words, 
the network takes a concept $c$ and the generated attention map {\small $L(\Mat{F},c)$} as input, 
and outputs how confident it feels that the concept exists in the frame.

%\CHK{can we change the energy function notation, this is getting crazy in the next few sections.
%\eg~\eqn~6. 
%Maybe something like this {\small $E_\texttt{net}(\Mat{F},c)$}.
%Reason: the c is always repeat\ldots}

To achieve this,
we first flatten the attention map {\small $L(\Mat{F},c)$} of size {\small $w \times h$} into a vector {\small $\Vec{V}_L \in \mathbb{R}^{w \cdot h}$},
and we employ the skip-gram model of word2vec~\cite{Mikolov_NIPS_2013} to convert a concept $c$ into a word embedding vector {\small $\Vec{V}_c\in\mathbb{R}^{c}$}.
We use the word vectors provided by \cite{Jain_ICCV_2015}, 
where words within a multi-words concept (\eg~{\it walking, with, dog}) are aggregated using Fisher Vectors.  
This word embedding can capture the semantic relatedness and compositionality between concepts, 
and lead to better performance compared with a one-hot embedding.
%\NOTE{is it necessary to show the results for word2vec and one-hot?}

We then embed {\small $\Vec{V}_L$} and {\small $\Vec{V}_c$} into a d-dimensional space using a two-layer fully-connected network with \textit{ReLU} nonlinearity between the layers.
The embedding is represented by \mbox{\small $f(\Vec{V}_{L},\Vec{V}_{c})\in\mathbb{R}^{d}$}.

The energy function is defined as,
\begin{equation}
	E_\texttt{net}(\Mat{F},c) = \Vec{W}_{f} f(\Vec{V}_{L},\Vec{V}_{c}),
	%E(L(\Mat{F},c),c) = \Vec{W}_{f} f(\Vec{V}_{L},\Vec{V}_{c}),
\end{equation}

\noindent
where {\small $\Vec{W}_f\in\mathbb{R}^{1 \times d}$} is the weight of the last fc layer for the Energy Network.
Similarly as \eqn~\ref{eqn:unsup_score}, the score for a video over each concept $c_i$ can be computed as,
\begin{equation}
	score(\Mat{F},c_i) = \dfrac{1}{N}\sum\nolimits_\Mat{F}{E_\texttt{net}(\Mat{F},c_i)}, %{E(L(\Mat{F},c_i),c_i)},
\label{eqn:sup_score}
\end{equation}

\noindent
and the predicted concept is the one with the highest score.

The proposed EnergyNet jointly optimizes two loss functions, energy loss and triplet loss.
The conceptual example of the optimization structure is shown in \fig~\ref{fig:network}.

\subsubsection{Energy Loss}

For each frame $\Mat{F}$ in the training videos, 
we denote the attention map generated by the ground truth concept  {\small $L(\Mat{F},c_{\texttt{gt}}^\Mat{F})$} as the \textit{true map}, 
and the attention map generated by any other concept {\small $L(\Mat{F},c_{\texttt{fa}}^\Mat{F})$} as the \textit{false map}.
Intuitively,
we want the true map to have higher energy than the false maps.
Therefore, we design the energy loss to be the hinge loss between the true map and false maps as,
\begin{equation}
	\resizebox{0.90\linewidth}{!}{$
	\begin{split}
	ELoss(\Mat{F},c_{\texttt{gt}}^\Mat{F},c_{\texttt{fa}}^\Mat{F}) = 
	                         & \max\{0, \; E_\texttt{net}(\Mat{F},c_{\texttt{fa}}^\Mat{F}) - 
                                       E_\texttt{net}(\Mat{F},c_{\texttt{gt}}^\Mat{F}) + m\}, \\
	                         & \forall \;\; c_{\texttt{fa}}^\Mat{F} \in {C}, \;\; c_{\texttt{fa}}^\Mat{F} \neq {c_{\texttt{gt}}^\Mat{F}},
	\end{split}
	$}	
\end{equation}

%\begin{equation}
%	\resizebox{0.90\linewidth}{!}{$
%	\begin{split}
%	ELoss(\Mat{F},c_{\texttt{gt}}^\Mat{F},c_{\texttt{fa}}^\Mat{F}) = 
%	                         & \max\{0,E(L(\Mat{F},c_{\texttt{fa}}^\Mat{F}),c_{\texttt{fa}}^\Mat{F}) - 
%                                       E(L(\Mat{F},c_{\texttt{gt}}^\Mat{F}),c_{\texttt{gt}}^\Mat{F}) + m\}, \\
%	                         & \forall \;\; c_{\texttt{fa}}^\Mat{F} \in {C}, \;\; c_{\texttt{fa}}^\Mat{F} \neq {c_{\texttt{gt}}^\Mat{F}},
%	\end{split}
%	$}	
%\end{equation}

\noindent
where $m$ is a margin enforced between true and false pairs.
Based on preliminary experiments, 
we set $m$ to 1 in all of our experiments.

\subsubsection{Hard Negative Mining}
\label{sec:hard_mining}

Generating all true-false concept pairs {\ $(c_{\texttt{gt}}^\Mat{F},c_{\texttt{fa}}^\Mat{F})$} with brute force approach would result in many pairs that easily satisfy 
{\small $E_\texttt{net}(\Mat{F},c_{\texttt{gt}}^\Mat{F})-E_\texttt{net}(\Mat{F},c_{\texttt{fa}}^\Mat{F}) \ge m$}, 
%{\small $E(L(\Mat{F},c_{\texttt{gt}}^\Mat{F}))-E(L(\Mat{F},c_{\texttt{fa}}^\Mat{F})) \ge m$}, 
thus having minimum contribution to the training process and can lead to slow convergence.
Therefore,
it is important to select the hard false concepts {\small $c_{\texttt{fa}}^\Mat{F}$} such that the energy loss is larger for those concepts that give 
{\small $\argmin_{c_{\texttt{fa}}^\Mat{F}} (E_\texttt{net}(\Mat{F},c_{\texttt{gt}}^\Mat{F})-E_\texttt{net}(\Mat{F},c_{\texttt{fa}}^\Mat{F}))$}.
%{\small $\argmin_{c_{\texttt{fa}}^\Mat{F}} (E(L(\Mat{F},c_{\texttt{gt}}^\Mat{F}))-E(L(\Mat{F},c_{\texttt{fa}}^\Mat{F})))$}.

To achieve this,
we mine the hard negative samples using an online approach.
We first generate large mini-batches where the false concepts are chosen randomly,
then we forward the mini-batches through the EnergyNet and select the top $K$ samples with the highest energy loss and
apply back-propagation on the selected $K$ samples.
To prevent early convergence to local minima in training stage, 
we generate smaller mini-batches at the start so that the initial $K$ samples are \textit{semi-hard}.
After the training loss decreases below a threshold, 
we increase the mini-batch size to raise the probability to generate stronger negative samples.
In addition to the $K$ hard samples, 
we also insert a few random samples into each training batch.

\begin{figure}[!t]
  \centerline{\includegraphics[width=1.0\columnwidth]{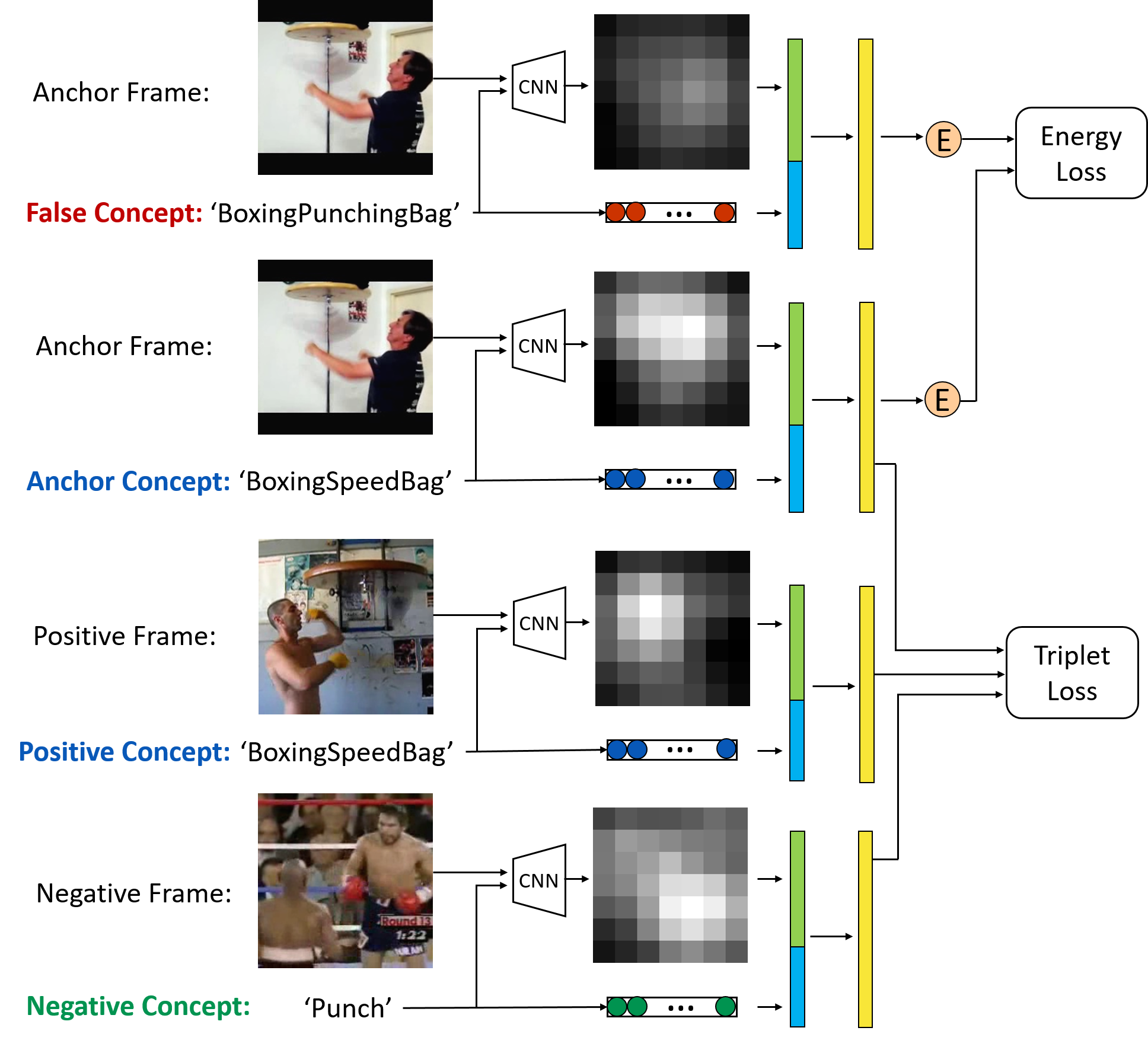}}	
  \vspace{-2ex}
  \caption
  	{
  	\small
		Illustration of Siamese EnergyNet with shared parameters.
		From top to bottom,  
		the inputs are \small $(\Mat{F},c_{\texttt{fa}}^{\Mat{F}})$, \small$(\Mat{F},c_{\texttt{gt}}^{\Mat{F}})$, 
		\small$(\Mat{F}^+,c^+)$ and 
		\small$(\Mat{F}^-,c^-)$.
		The energy loss and triplet loss are jointly optimized.
  	}
  \label{fig:network}
\end{figure}

\subsubsection{Triplet Loss on Embedding}

Inspired by~\cite{Wang_ICCV_2015,Schroff_CVPR_2015,Hoffer_SIMBAD_2015},
we construct a triplet loss on the embedding space to learn a more representative feature embedding {\small $f(\Vec{V}_{L},\Vec{V}_{c})$} 
that capture intra-class similarity and inter-class difference of the \textit{true maps}. 
We use {\small $f(\Mat{F},c)$} to denote {\small $f(\Vec{V}_{L},\Vec{V}_{c})$} in this subsection for simplicity purposes.
%\TODO{the notation of $f()$ is inconsistent from \eqn~4, I leave it to you to consolidate it. Maybe somewhere in above sentence you need to indicate we use xxxxxx to denote yyyyy in this subsection for simplicity purposes.}

Given two frames {\small $\Mat{F}_1$}, {\small $\Mat{F}_2$}, and their respective true concepts {\small $c_{\texttt{gt}}^{\Mat{F}_1}$}, {\small $c_{\texttt{gt}}^{\Mat{F}_2}$},
we define the distance in the embedding space {\small $f(\Mat{F},c)$} based on the Cosine distance,
\begin{equation}
	D\big(f(\Mat{F}_1,c_{\texttt{gt}}^{\Mat{F}_1}),f(\Mat{F}_2,c_{\texttt{gt}}^{\Mat{F}_2})\big) = 1 - \frac{f(\Mat{F}_1,c_{\texttt{gt}}^{\Mat{F}_1})\cdot{f(\Mat{F}_2,c_{\texttt{gt}}^{\Mat{F}_2})}}{\lVert f(\Mat{F}_1,c_{\texttt{gt}}^{\Mat{F}_1}) \rVert \lVert f(\Mat{F}_2,c_{\texttt{gt}}^{\Mat{F}_2}) \rVert}.
\end{equation}

We design our triplet loss on {\small $f(\cdot)$},
such that the distance between embeddings of the same class is small,
and the distance between embeddings of different classes is larger.
We select an anchor frame {\small $\Mat{F}$}, a positive frame {\small $\Mat{F}^+$} and a negative frame {\small $\Mat{F}^-$} from three videos {\small $V$}, {\small $V^+$}, {\small $V^-$},
where {\small $V$} and {\small $V^+$} are of the same class, and {\small $V^-$} is from a different class.
%We select an anchor map, a positive map and a negative map from three frames $\Mat{F}_{\texttt{anc}}$, $\Mat{F}_{\texttt{pos}}$, $\Mat{F}_{\texttt{neg}}$,
%where the frames are randomly selected from three different videos $V_{\texttt{anc}}$, $V_{\texttt{pos}}$, $V_{\texttt{neg}}$.
We generate the \textit{anchor map}, the \textit{positive map} and the \textit{negative map} as the \textit{true maps} of the three frames,
\begin{equation}
	\begin{split}
	L\;\; &= L(\Mat{F},c), \;\;\;\;\; \forall \; c\;\; = c_{\texttt{gt}}^{\Mat{F}},\\
	L^+ &= L(\Mat{F}^+,c^+), \; \forall \; c^+ = c, c^+ = c_{\texttt{gt}}^{\Mat{F}^+},\\
	L^- &= L(\Mat{F}^-,c^-), \; \forall \; c^- \neq c, c^- = c_{\texttt{gt}}^{\Mat{F}^-}.\\		
	\end{split}
\end{equation}

%\begin{equation}
%	\begin{split}
%	L_{\texttt{anc}} &= L(\Mat{F}_{\texttt{anc}},c_{\texttt{anc}}), \; \forall \; c_{\texttt{anc}} = c_{\texttt{gt}}^{\Mat{F}_{\texttt{anc}}},\\
%	L_{\texttt{pos}} &= L(\Mat{F}_{\texttt{pos}},c_{\texttt{pos}}), \; \forall \; c_{\texttt{pos}} = c_{\texttt{anc}}, c_{\texttt{pos}} = c_{\texttt{gt}}^{\Mat{F}_{\texttt{pos}}},\\
%	L_{\texttt{neg}} &= L(\Mat{F}_{\texttt{neg}},c_{\texttt{neg}}), \; \forall \; c_{\texttt{neg}} \neq c_{\texttt{anc}}, c_{\texttt{neg}} = c_{\texttt{gt}}^{\Mat{F}_{\texttt{neg}}}.\\		
%	\end{split}
%\end{equation}

\noindent
The embeddings {\small $f(\Mat{F},c)$} for the three frames are then computed by forwarding $L$ and $c$ through the embedding layers.
%We generate the \textit{anchor map} as the \textit{true map} for $F_a$: $L_a = L(F_a,c_a)$, where $c_a = c_{gt}^{F_a}$.
%We generate the \textit{positive map} as the \textit{true map} for $F_p$: $L_p = L(F_p,c_p)$,
%where $F_p$ is from the same class as $F_a (c_p = c_a,c_p = c_{gt}^{F_p})$.
%And we generate the \textit{negative map} as the \textit{true map} for $F_n$:
%$L_n = L(F_n,c_n)$,
%where $F_n$ is from a different class with $F_a (c_n \neq {c_a}, c_n = c_{gt}^{F_n})$.
%\NOTE{is it better to change the above 3 sentence into equation using split? \ie~3 equations. Then below the equation say xxx and xxx denotes the true map of achor map and positive map, respectively, and xxx denotes the negative map.}

Here, we enforce the following condition 
\begin{equation}
	D\big(f(\Mat{F},c),f(\Mat{F}^-,c^-)\big) > D\big(f(\Mat{F},c),f(\Mat{F}^+,c^+)\big),
\end{equation}

\noindent
by defining the triplet loss as,
\begin{equation}
	\resizebox{0.9\linewidth}{!}{$
	\begin{split}
	TLoss&(\Mat{F},\Mat{F}^+,\Mat{F}^-,c,c^+,c^-) = \\ 
	     &\max\big\{0, \; D\big(f(\Mat{F},c),f(\Mat{F}^+,c^+)\big) - D\big(f(\Mat{F},c),f(\Mat{F}^-,c^-)\big)+\beta\big\},
	\end{split}
	$}
\end{equation}

%\begin{equation}
%	\resizebox{0.9\linewidth}{!}{$
%	\begin{split}
%	TLoss&(\Mat{F}_{\texttt{anc}},\Mat{F}_{\texttt{pos}},\Mat{F}_{\texttt{neg}},c_{\texttt{anc}},c_{\texttt{pos}},c_{\texttt{neg}}) = \\ 
%	     &\max\{0,D\big(f(\Mat{F}_{\texttt{anc}},c_{\texttt{anc}}),f(\Mat{F}_{\texttt{pos}},c_{\texttt{pos}})\big) - D\big(f(\Mat{F}_{\texttt{anc}},c_{\texttt{anc}}),f(\Mat{F}_{\texttt{neg}},c_{\texttt{neg}})\big)+\beta\},
%	\end{split}
%	$}
%\end{equation}

\noindent
where $\beta$ represents the margin, and is set to 0.5 in the experiment based on preliminary experiments.

\subsubsection{Joint Learning}

During the training phase, we want to jointly optimize the energy loss and the triplet loss.
To achieve this, we create a Siamese EnergyNet (see \fig~\ref{fig:network}) comprised of 4 instances of the same EnergyNet (with shared parameters).
The input to the Siamese network is {\small $\big\{(\Mat{F},c_{fa}^{\Mat{F}}),(\Mat{F},c_{gt}^{\Mat{F}}),(\Mat{F}^+,c^+),(\Mat{F}^-,c^-)\big\}$}.
%\begin{equation}
%{\small \big\{(\Mat{F},c_{fa}^{\Mat{F}}),(\Mat{F},c_{gt}^{\Mat{F}}),(\Mat{F}^+,c^+),(\Mat{F}^-,c^-)\big\}}.
%\end{equation}

\noindent
In practice, we pre-compute the attention maps $L(F,c)$ for all $F$ and $c$,
so that training can be much faster.

Since the input consists of both \textit{false anchor map} {\small $L(\Mat{F},c_{fa}^{\Mat{F}})$} and \textit{negative true map} {\small $L(\Mat{F}^-,c^-)$}, 
we want to select the hard negatives {\small $\big\{(\Mat{F},c_{fa}^{\Mat{F}}),(\Mat{F}^-,c^-)\big\}$} that maximize both the energy loss and the triplet loss.
We mine for the hard negatives using the same online approach as Section~\ref{sec:hard_mining}.
As recommended in practice by~\cite{Schroff_CVPR_2015,Wang_ICCV_2015},
we do not select hard positives.
We apply Stochastic Gradient Descent (SGD) to train the network.
We let $d=64$ for the embedding {\small $f(\cdot)\in\mathbb{R}^{d}$}.
The learning rate is set as 0.0001, and we use a weight decay of 0.0005.
 
\section{Experiments}
\label{sec:experiments}

\subsection{Datasets}
\label{sec:dataset}
We consider two labeled video datasets for two tasks: human interaction recognition and action recognition.
Both datasets contain videos in unconstrained environments and have been widely used for benchmarking action/interaction recognition methods.
For each video dataset, we have a corresponding Web image dataset.

% \noindent\textbf{Human Interaction Recognition}.
\subsubsection{Human Interaction Recognition}
We use the TV Human Interaction (TVHI)~\cite{Perez_PAMI_2012} dataset for video recognition.
It consists of 300 video clips compiled from 23 different TV shows.
It contains four types of human interactions: Handshake, Highfive, Hug and Kiss.
Each interaction has 50 videos, and the remaining 100 videos are negative examples.
We use the 200 positive videos, and keep the train/test split as~\cite{Perez_PAMI_2012}.
The dataset is generally considered challenging due to occlusion and viewpoint changes.

We collected an image dataset corresponding to the four types of interactions, 
namely the Human Interaction Image (HII) dataset\footnote{available via \url{https://doi.org/10.5281/zenodo.832380}}.
Given an action name, 
we crawled Web images from Commercial Search Engines (Google, Bing and Flickr) using keyword search.
Duplicate images were removed by comparing their color histogram.
In total, we collected 17.5K images. 
Since the images are noisy, we manually filtered out the irrelevant ones that do not contain a concept.
The filtered dataset contains 2410 images with at least 550 images per class.
For experiments, we evaluate with both the noisy data and the filtered one.

% \noindent\textbf{Action Recognition}.
\subsubsection{Action Recognition}
We use UCF101~\cite{Soomro_CORR_2012}, a large-scale video dataset for action recognition.
It consists of 101 action classes, over 13k clips and 27 hours of video collected from YouTube.
The videos are captured under various lighting conditions with camera motion, occlusion and low frame quality, making the task challenging.
We use the three provided train/test split, and report the classification accuracy for evaluation.

We use BU101~\cite{Ma_PR_2017} as the Web image dataset,
which has class-to-class correspondence with the UCF101 dataset.
It was collected from the Web using key phrases, and then manually filtered to remove irrelevant frames.
It comprises 23.8K images with a minimum of 100 images per class.

\subsection{Training CNNs with Web Images}
\label{sec:train_img}

CNNs pre-trained from ImageNet have been widely used for action recognition~\cite{Simonyan_NIPS_2014,Gan_CVPR_2016}.
We choose the state-of-the-art model (\ie~101-layer ResNet~\cite{He_CVPR_2016}) and fine-tune it on the Web image dataset.
To show that our method can generalize to other CNN architectures, 
we also evaluate VGGNet16~\cite{Simonyan_CORR_2014} for human interaction recognition.

To pre-process the images, 
we first resize the shorter side to 224 pixels while keeping the aspect ratio.
Then we apply center crop to obtain the 224$\times$224 input compatible with the CNN architecture.
We augment the training images with random horizontal flipping.
We apply SGD with a mini-batch size of 32.
For both HII dataset and BU101 dataset, we randomly split 30\% of the images for validation.
The result on the validation set is shown in~\tab~\ref{tbl:imagecnn}.
The CNN models trained on HII-noisy obtain better performance than HII-filter.
This finding is consistent with~\cite{Joulin_ECCV_2016,Krause_ECCV_2016},
where large-scale noisily labeled images can be effective for image classification task.
In addition, 
model trained on HII has higher accuracy than BU101.
This is because it contains fewer classes and more images per class.

The Web-image trained CNN is then used to generate spatial attention maps for video frames.
The size of the attention map for ResNet and VGGNet is $7\times7$ and $14\times14$, respectively.
%In this paper, 
%we adapt two widely-used CNN models, 
%namely ResNet~\cite{He_CVPR_2016} and VGGNet~\cite{Simonyan_CORR_2014},
%where the size of the attention map is fixed at $7\times7$ and $14\times14$, 
%respectively.

\begin{table}[!t]
	\centering
	\caption
		{
		\small	
		Classification accuracy (\%) on validation set for training CNNs with Web images.
		}
	\label{tbl:imagecnn}
	\vspace{-2ex}
	\begin{tabular}{l|c|c|c} 
		\toprule
		CNN Model\hspace{2ex} & HII-noisy & HII-filtered & BU101		\\
		\midrule
		ResNet101 & 95.2 & 94.1 & 88.3  \\
		VGGNet16 & 90.1 & 89.4 & -  \\
		\bottomrule
	\end{tabular}
\end{table}

\subsection{Experiment Setup}

\noindent\textbf{Unsupervised Domain Adaptation}. 
In this scenario, 
we directly apply the Web images trained CNNs to classify the videos.
We examine two types of classifiers:
\begin{enumerate}[leftmargin=0.4cm]
	\item[--]
	{\bf CNN}: 
	We use the output score from the last fc layer of the CNN to classify each frame.
	The class of the video is determined by voting of the frames' classes.
	We select the majority frames that has the same class, 
	and apply late fusion (average) on the frame-level scores to calculate video-level score.
	We also experiment with averaging all frame-level scores, where the performance are slightly degrade.
	\item[--]
	{\bf Unsupervised Attention (UnAtt)}: 
	Proposed method delineated in Section~\ref{sec:uda}, 
	where a sliding-window approach is used to compute energy from the attention maps.
	The energy score is computed with~\eqn~\ref{eqn:unsup_score}.
\end{enumerate}	
\noindent\textbf{Supervised Domain Adaptation}. 
In this setting, 
we further train the image-trained classifiers on the labeled training videos with four methods:
\begin{enumerate}[leftmargin=0.4cm]
	\item[--]
	{\bf SVM}: 
	We extract features from the penultimate layer of the image-trained CNN (pool5 for ResNet and fc7 for VGG), and train one-versus-rest linear SVM classifiers with the soft margin cost set as 1. 
	The video class is predicted with majority voting.
	%	We use the LibSVM toolbox~\cite{Chang_LIBSVM_2011} that can produce probability estimates for computing mAP.
	\item[--]
	{\bf finetune+CNN}: 
	We fine-tune the CNNs on frames from the training videos.
	The training process is described in Section~\ref{sec:train_img}.
	Then we use the output score from the last layer of the CNN to classify given videos, where voting is applied.
	\item[--]
	{\bf finetune+UnAtt}:
	We apply the Unsupervised Attention method with the fine-tuned CNN.	
	\item[--]
	{\bf finetune+EnergyNet}: 
	We train an EnergyNet (Section~\ref{sec:sda}) using spatial attention maps generated by the fine-tuned CNN, and calculate the score with~\eqn\ref{eqn:sup_score}. 
\end{enumerate}

\begin{table}[!t]
	\centering
	\caption
		{
		\small	
		Mean Average Precision (mAP) (\%) for both Unsupervised (shaded in blue) and Supervised Domain Adaptation on TVHI dataset. 
		CNNs pre-trained on both the filtered images and noisy images are evaluated.
		}
	\label{tbl:interaction}
	\vspace{-2ex}
	\resizebox{1.0\linewidth}{!}{
	\begin{tabular}{l|l|c|c} 
		\toprule
		CNN Model 					& Method 		& HII-filtered 	& HII-noisy			\\ \midrule
		\multirow{6}{*}{ResNet101}	& \CellB{CNN} 	& \CellB{89.7} 	& \CellB{91.6}		\\
									& \CellB{UnAtt} & \CellB{94.3}  & \CellB{\bf{96.0}}	\\
		%&SVM & -  & 78 & - & 80\\
									& finetune+CNN & 92.6  & 92.9 \\
									& finetune+UnAtt & 94.7 & 96.3 \\
									& finetune+EnergyNet& 96.8 & \bf{98.7} \\ \midrule
		\multirow{6}{*}{VGG16}		& \CellB{CNN} 	& \CellB{86.7}  	&\CellB{85.3}  \\
									& \CellB{UnAtt} 	& \CellB{\bf{89.0}} & \CellB{86.9}\\	
		%&SVM & -  & 74 & - & 71\\
									& finetune+CNN		& 89.5  & 88.9  \\
									& finetune+UnAtt 	& 90.5	& 89.4\\
									& finetune+EnergyNet	& \bf{91.7}  & 90.7  \\		
		\bottomrule
	\end{tabular}
	}
\end{table}

%\begin{table}[!t]
%	\centering
%	\caption
%	{
%		\small	
%		Mean Average Precision (mAP) and classification accuracy (\%) for both Unsupervised and Supervised Domain Adaptation on TVHI dataset.
%	}
%	\label{tbl:interaction}
%	\vspace{-2ex}
%	\resizebox{1.0\linewidth}{!}{
%		\begin{tabular}{l|l|c|c||c|c} 
%			\toprule
%			\multirow{3}{*}{CNN Model} & \multirow{3}{*}{Method} &  \multicolumn{2}{c||}{HII-filtered} & \multicolumn{2}{c}{HII-noisy}\\
%			\cmidrule{3-6}
%			& & mAP & Acc &  mAP & Acc \\
%			\midrule
%			\multirow{6}{*}{ResNet101}
%			&CNN & 89.7 & 81 & 91.6 & 82\\
%			&UnAtt & 94.3 & 90 & \bf{96.0} & \bf{91}\\
%			&SVM & -  & 78 & - & 80\\
%			&finetune+CNN & 92.6 & 83 & 92.9 & 84\\
%			&finetune+UnAtt & 94.7 & 90 & 96.3 & 92\\
%			&finetune+EnergyNet& 96.8 & 93 & \bf{98.7} & \bf{94}\\
%			\midrule
%			\multirow{6}{*}{VGG16}
%			&CNN & 86.7 & 75 &85.3 & 74 \\
%			&UnAtt & \bf{89.0} & \bf{79} & 86.9 & 77  \\	
%			&SVM & -  & 74 & - & 71\\
%			&finetune+CNN& 89.5 & 81 &88.9 & 80 \\
%			&finetune+UnAtt & 90.5 & 82 & 89.4 & 81\\
%			&finetune+EnergyNet& \bf{91.7} & \bf{84} & 90.7 & 83  \\		
%			\bottomrule
%		\end{tabular}
%	}
%\end{table}	
\begin{table}[!t]
	\centering
	\caption
		{
		\small	
		Comparison with state-of-the-art methods on TVHI dataset.
		}
	\label{tbl:interaction_state}
	\vspace{-2ex}
	\begin{tabular}{l|c} 
		\toprule
		Methods\hspace{22ex} & mAP \\
		\midrule
		Patron~\etal~\cite{Perez_PAMI_2012} & 42.4\\
		Hoai~\etal~\cite{Hoai_ACCV_2014} & 71.1\\
		Wang~\etal~\cite{Wang_MM_2015}  & 78.2\\
		NoImage+ResNet & 44.9\\
		ResNet+UnAtt & 96.0\\
		ResNet+finetune+EnergyNet & \bf{98.7}\\
		
		\bottomrule
	\end{tabular}
\end{table}	

\subsection{Human Interaction Recognition Task}
\label{sec:interaction_experiment}

We use the CNNs pre-trained on Human Interaction Image (HII) dataset, and evaluated the methods on the test set of TVHI dataset, which consists a total of 100 videos with 25 videos per class.
We report the mean average precision (mAP) for evaluation purpose.
For supervised methods, we train on all frames from the 100 training videos of TVHI. 
To test the scalability of our methods,
we investigate with using both noisy and filtered images to pre-train the CNNs.

\noindent\textbf{Results}. 
\tab~\ref{tbl:interaction} shows the results for both unsupervised and supervised domain adaptation,
The proposed methods (UnAtt and EnergyNet) outperforms corresponding baselines using either ResNet101 or VGG16.
The proposed unsupervised method (UnAtt) can achieve better performance compared with the supervised baseline method (\ie~finetune+CNN with ResNet101),
demonstrating its efficacy to transfer knowledge across domains.

By fine-tuning on the video frames, both the baseline methods and the proposed method achieve performance improvement.
For ResNet101, using finetune+EnergyNet leads to $+5.8$ in mAP when compared with finetune+CNN,
and $+2.4$ against finetune+UnAtt.
This shows that the EnergyNet can further adapt to the video domain by training on the attention maps.

A surprising result is that ResNet101 can benefit from training on large amount of noisy images as compared with using filtered images.
This demonstrates the robustness of ResNet in learning good feature representations from noisy data.
VGG16 only degrades by a small margin using noisy images instead of filtered ones.
This indicates that our method can directly make use of weakly-labeled Web images and does not require manual labeling.

\noindent\textbf{Comparison with state-of-the-art}. 
We compare our method with the state-of-the-art approaches for interaction recognition.
Since all previous approaches do not use deep learning, we create another baseline NoImage+ResNet,
where we directly fine-tune a ResNet101 model pre-trained from ImageNet on the video frames (without Web images).
As shown in \tab~\ref{tbl:interaction_state},
we can see that using Web images as auxiliary training data significantly improves the performance,
especially when the amount of the available training video is low.

\subsection{Action Recognition Task}

We use the CNN model pre-trained on BU101 dataset, and evaluate the methods on the test set of UCF101 dataset.
We report the classification accuracy averaged over the three test splits.
To reduce redundant frames and training time, 
we sample one frame for every five from the videos for both training and test.
Based on the experimental results on TVHI,
we choose ResNet101 as the CNN model for this task.

\begin{table}[!t]
	\centering
	\caption
		{
		\small	
		Mean classification accuracy of Unsupervised Domain Adaptation on UCF101 (averaged over three test splits).
		}
	\label{tbl:action_unsup}
	\vspace{-2ex}
	\begin{tabular}{l|c|c} 
		\toprule
		Method\hspace{5ex} & Top-1     & Top-3     \\ \midrule
		CNN   & 62.5      & 78.5      \\
		UnAtt & \bf{66.4} & \bf{82.4} \\
		\bottomrule
	\end{tabular}
\end{table}	

\begin{figure}[!t]
 \centering
  %\begin{minipage}{1.0\columnwidth}
  	\begin{subfigure}{0.49\columnwidth}  		
		 \centerline{\includegraphics[width=\linewidth]{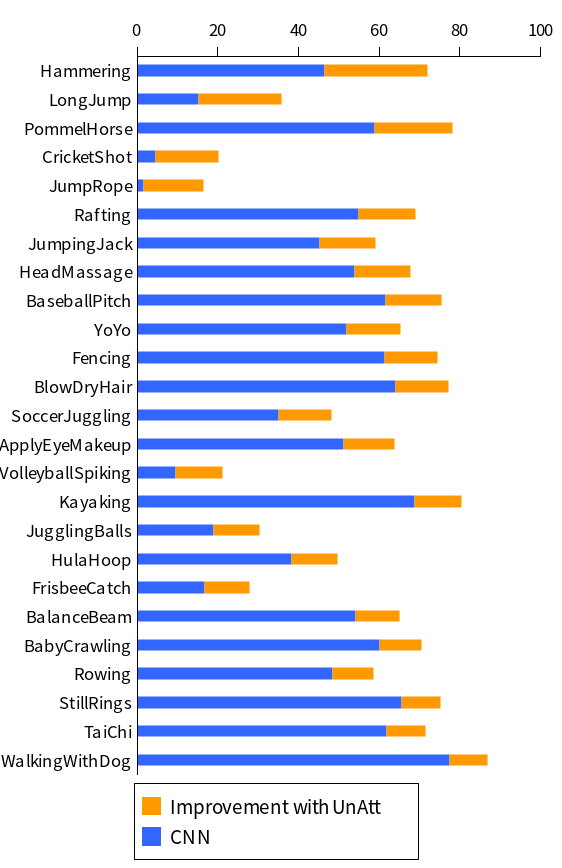}}
		 \subcaption{\footnotesize Unsupervised Domain Adaption}
		 \label{fig:sub:unsup_improve}
  	\end{subfigure}
  	%\hfill
  	\begin{subfigure}{0.49\columnwidth}
  		\centerline{\includegraphics[width=\linewidth]{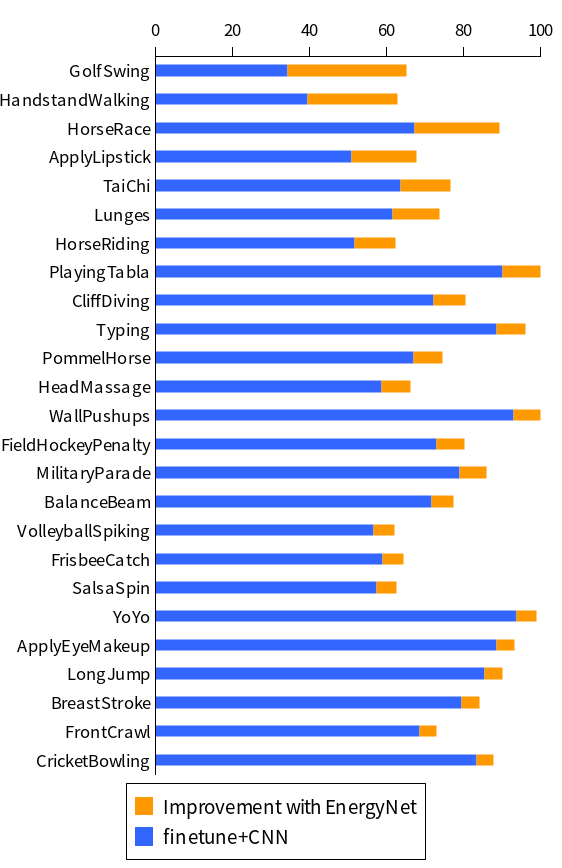}}
  		\subcaption{\footnotesize Supervised Domain Adaption}
  		\vspace{-1ex}
  	\end{subfigure}
  %\end{minipage}
	\vspace{-2ex}
  \caption
    {
   	\small
		The 25 action classes with the largest accuracy improvement on UCF101 dataset.
		Supervised domain adaptation is performed with 20\% training videos.
    } 
  \vspace{-2ex}
  \label{fig:improve}
\end{figure}

\noindent\textbf{Results}.
\tab~\ref{tbl:action_unsup} shows the result for unsupervised domain adaptation,
where the efficacy of the proposed method (UnAtt) is proved.
\fig~\ref{fig:improve} shows the 25 action classes with the largest accuracy improvement, for both unsupervised and supervised scenario.

\begin{figure}[!t]
	\centerline{\includegraphics[width=1.0\columnwidth]{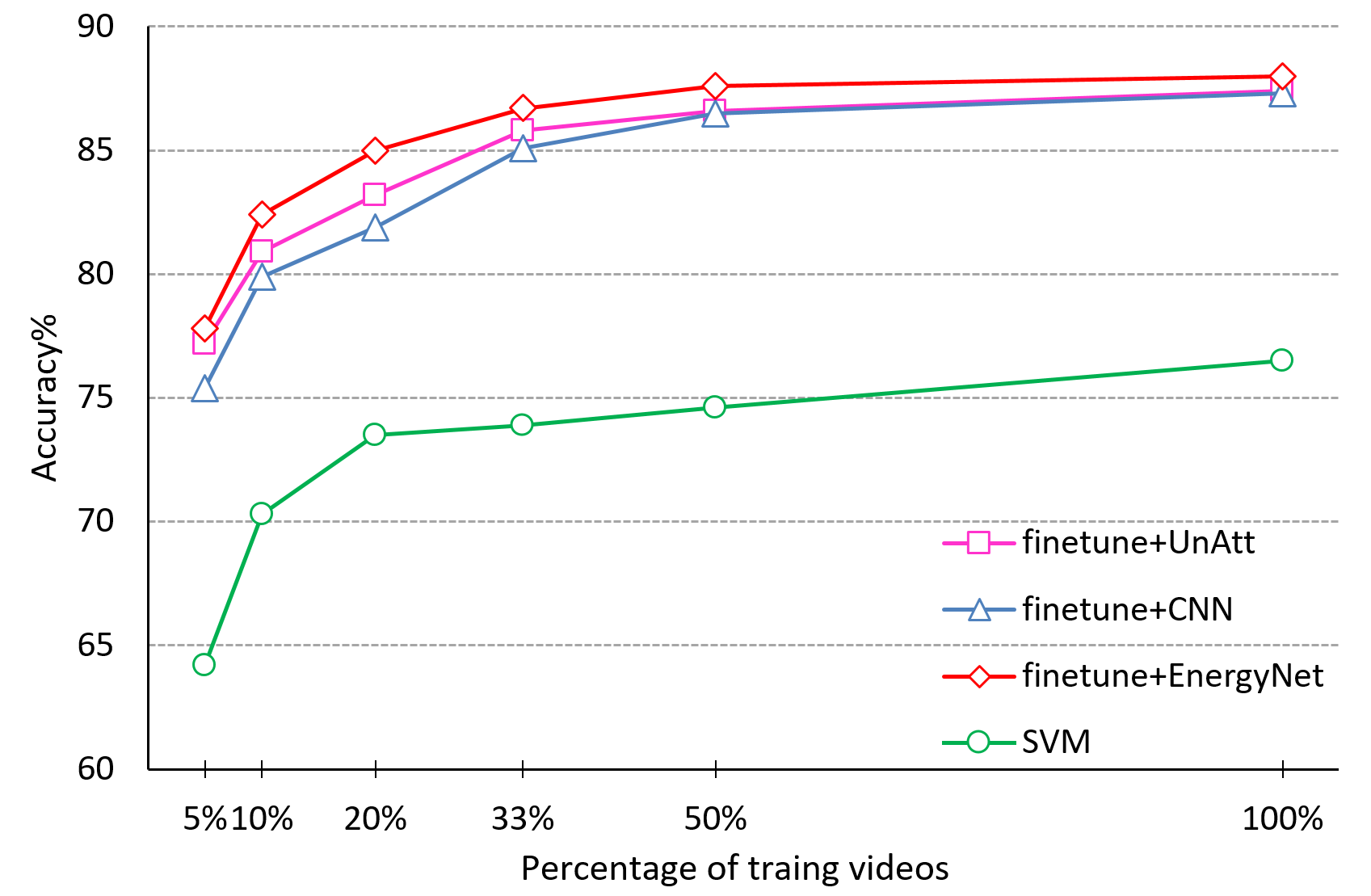}}	
	\vspace{-2ex}
	\caption
	{
		\small
		Mean classification accuracy of Supervised Domain Adaptation with different number of training videos used.
	}
	\label{fig:action_sup}
\end{figure}

% \noindent\textbf{Vary the Number of Training Videos}.
For supervised domain adaptation,
we study how the amount of training data in the target domain influence the performance of the proposed method.
Each training split in the UCF101 dataset comprises around 95 clips per action class.
We randomly sample 5\%, 10\%, 20\%, 33\%, 50\% and 100\% videos from the training set,
and report the performance for each sampled set.
The experiments are separately done on the three train/test splits, and the result is averaged across three test splits and reported in \fig~\ref{fig:action_sup}.

\begin{table}[!t]
	\centering
	\caption
		{
		\small	
		Comparison with state-of-the-art methods on UCF101. * refers to methods that use Web data.
		}
	\label{tbl:action_state}
	\vspace{-2ex}
	\begin{tabular}{l|l|c} 
		\toprule
		&Method & Accuracy (\%) \\
		\midrule
		\multirow{5}{*}{W/O Motion}
		&Spatial stream network~\cite{Simonyan_NIPS_2014}&73.0\\
		&LRCN~\cite{Donahue_CVPR_2015} & 71.1\\
		&Karpathy~\etal~\cite{Karpathy_CVPR_2014} & 65.4\\
		&* Webly-supervised~\cite{Gan_ECCV_2016}& 69.3\\
		&* Ma~\etal spatial~\cite{Ma_PR_2017}&83.5\\
		
		\hline
		\multirow{6}{*}{With Motion}
		&Two-stream network~\cite{Simonyan_NIPS_2014}& 88.0 \\
		&IDT+FV~\cite{Wang_ICCV_2013} & 87.9\\
		&C3D~\cite{Tran_ICCV_2015}  & 82.3\\
		&TDDs~\cite{Wang_CVPR_2015} & 90.3 \\
		&TDDs+IDT-FV~\cite{Wang_CVPR_2015} & 91.5 \\
		&* Ma~\etal spatial+IDT-FV~\cite{Ma_PR_2017} & 91.1 \\
		\hline
		\multirow{3}{*}{Ours}
		&* UnAtt (no training video)& 66.4\\
		%&EnergyNet (10\% training video)& 82.4\\
		&* EnergyNet (20\% training video)& 85.0 \\
		&* EnergyNet (all training video)& 88.0\\
		
		\bottomrule
	\end{tabular}
	\vspace{-2ex}
\end{table}	

\begin{figure*}[!t]
	\captionsetup[subfigure]{labelformat=empty,justification=centering}
	\footnotesize
 \centering
  \begin{minipage}{1.0\columnwidth}
  	\begin{minipage}{0.3\columnwidth}
		 \centerline{\includegraphics[width=\linewidth]{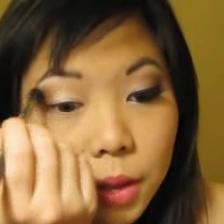}}
		 \subcaption{Apply Eye Makeup}
  	\end{minipage}
  	\hfill
  	\begin{minipage}{0.3\columnwidth}
  		\centerline{\includegraphics[width=\linewidth]{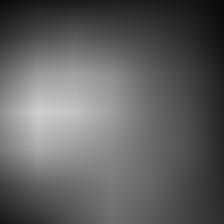}}
  		\subcaption{\textcolor{ForestGreen}{Apply Eye Makeup}}% \\ $Energy=22.3$}
  	\end{minipage}
  	\hfill
  	\begin{minipage}{0.3\columnwidth}
  		\centerline{\includegraphics[width=\linewidth]{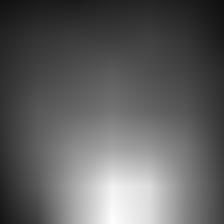}}
  		\subcaption{\textcolor{BrickRed}{Apply Lipstick}}% \\$Energy=21.1$}
  	\end{minipage}
  \end{minipage}
  \hfill
  \begin{minipage}{1.0\columnwidth}
  	\begin{minipage}{0.3\columnwidth}
  		\centerline{\includegraphics[width=\linewidth]{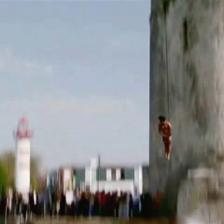}}
  		\subcaption{Cliff Diving}
  	\end{minipage}
  	\hfill
  	\begin{minipage}{0.3\columnwidth}
  		\centerline{\includegraphics[width=\linewidth]{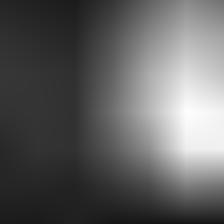}}
  		\subcaption{\textcolor{ForestGreen}{Cliff Diving}}% \\ $Energy=12.0$}
  	\end{minipage}
  	\hfill
  	\begin{minipage}{0.3\columnwidth}
  		\centerline{\includegraphics[width=\linewidth]{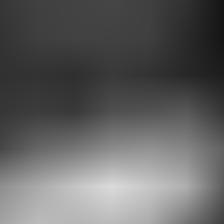}}
  		\subcaption{\textcolor{BrickRed}{Pole Vault}}% \\$Energy=9.7$}
  	\end{minipage}
  \end{minipage}
  \vspace{1ex}
  
  \begin{minipage}{1.0\columnwidth}
  	\begin{minipage}{0.3\columnwidth}
  		\centerline{\includegraphics[width=\linewidth]{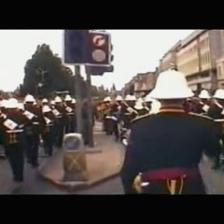}}
  		\subcaption{Band Marching}
  	\end{minipage}
  	\hfill
  	\begin{minipage}{0.3\columnwidth}
  		\centerline{\includegraphics[width=\linewidth]{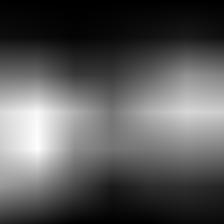}}
  		\subcaption{\textcolor{ForestGreen}{Band Marching}}% \\ $Energy=16.4$}
  	\end{minipage}
  	\hfill
  	\begin{minipage}{0.3\columnwidth}
  		\centerline{\includegraphics[width=\linewidth]{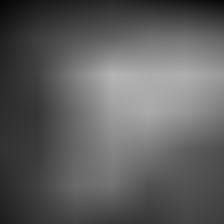}}
  		\subcaption{\textcolor{BrickRed}{Pizza Tossing}}% \\$Energy=12.6$}
  	\end{minipage}
  \end{minipage}
  \hfill
  \begin{minipage}{1.0\columnwidth}
  	\begin{minipage}{0.3\columnwidth}
  		\centerline{\includegraphics[width=\linewidth]{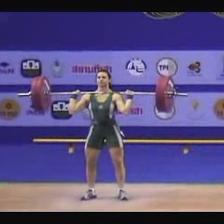}}
  		\subcaption{Clean And Jerk}
  	\end{minipage}
  	\hfill
  	\begin{minipage}{0.3\columnwidth}
  		\centerline{\includegraphics[width=\linewidth]{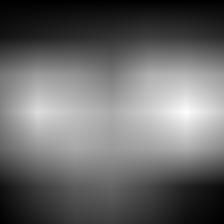}}
  		\subcaption{\textcolor{ForestGreen}{Clean And Jerk}}% \\ $Energy=17.5$}
  	\end{minipage}
  	\hfill
  	\begin{minipage}{0.3\columnwidth}
  		\centerline{\includegraphics[width=\linewidth]{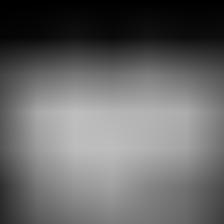}}
  		\subcaption{\textcolor{BrickRed}{Lunges}}% \\$Energy=15.8$}
  	\end{minipage}
  \end{minipage}
  \vspace{1ex}

  \begin{minipage}{1.0\columnwidth}
  	\begin{minipage}{0.3\columnwidth}
  		\centerline{\includegraphics[width=\linewidth]{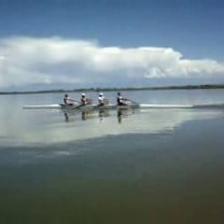}}
  		\subcaption{Rowing}
  	\end{minipage}
  	\hfill
  	\begin{minipage}{0.3\columnwidth}
  		\centerline{\includegraphics[width=\linewidth]{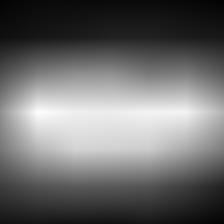}}
  		\subcaption{\textcolor{ForestGreen}{Rowing}}% \\ $Energy=20.8$}
  	\end{minipage}
  	\hfill
  	\begin{minipage}{0.3\columnwidth}
  		\centerline{\includegraphics[width=\linewidth]{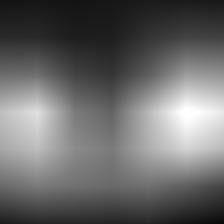}}
  		\subcaption{\textcolor{BrickRed}{Pole Vault}}% \\$Energy=20.0$}
  	\end{minipage}
  \end{minipage}
  \hfill
  \begin{minipage}{1.0\columnwidth}
  	\begin{minipage}{0.3\columnwidth}
  		\centerline{\includegraphics[width=\linewidth]{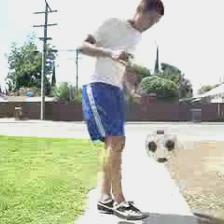}}
  		\subcaption{Soccer Juggling}
  	\end{minipage}
  	\hfill
  	\begin{minipage}{0.3\columnwidth}
  		\centerline{\includegraphics[width=\linewidth]{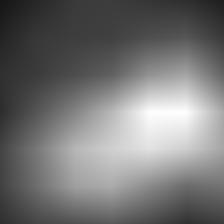}}
  		\subcaption{\textcolor{ForestGreen}{Soccer Juggling}}% \\ $Energy=14.1$}
  	\end{minipage}
  	\hfill
  	\begin{minipage}{0.3\columnwidth}
  		\centerline{\includegraphics[width=\linewidth]{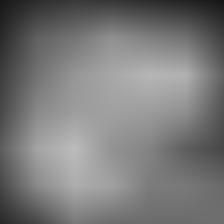}}
  		\subcaption{\textcolor{BrickRed}{Golf Swing}} %\\$Energy=12.7$}
  	\end{minipage}
  \end{minipage}
        
  \vspace{-2ex}
  \caption
    {
		\small
		Example frames from UCF101 and their corresponding spatial attention maps (resized to image size) for two concepts generated by the fine-tuned CNN. 
		The \textcolor{ForestGreen}{green} concept is the one correctly predicted by the proposed EnergyNet,
		while the \textcolor{BrickRed}{red} concept is the one wrongly predicted by the baseline CNN method.
		The EnergyNet correctly assigns a higher energy for the ground truth concept.
		%\NOTE{we have space budget, if it improve quality of this paper, we can add in more images. Not compulsary}
    } 
    \vspace{-2ex}
  \label{fig:visualization}
\end{figure*}

%  \begin{minipage}{1.0\columnwidth}
%  	\begin{minipage}{0.3\columnwidth}
%  		\centerline{\includegraphics[width=\linewidth]{visualize/BlowDryHair/img}}
%  		\subcaption{Blow Dry Hair\\ \hfill}
%  	\end{minipage}
%  	\hfill
%  	\begin{minipage}{0.3\columnwidth}
%  		\centerline{\includegraphics[width=\linewidth]{visualize/BlowDryHair/correct-13'1}}
%  		\subcaption{\textcolor{ForestGreen}{Blow Dry Hair} \\ $Energy=13.1$}
%  	\end{minipage}
%  	\hfill
%  	\begin{minipage}{0.3\columnwidth}
%  		\centerline{\includegraphics[width=\linewidth]{visualize/BlowDryHair/YoYo-11'7}}
%  		\subcaption{\textcolor{BrickRed}{YoYo} \\$Energy=11.7$}
%  	\end{minipage}
%  \end{minipage}
%  \vspace{1ex}

%  \begin{minipage}{1.0\columnwidth}
%  	\begin{minipage}{0.3\columnwidth}
%  		\centerline{\includegraphics[width=\linewidth]{visualize/BrushingTeeth/img}}
%  		\subcaption{Brushing Teeth\\ \hfill}
%  	\end{minipage}
%  	\hfill
%  	\begin{minipage}{0.3\columnwidth}
%  		\centerline{\includegraphics[width=\linewidth]{visualize/BrushingTeeth/correct-13'4}}
%  		\subcaption{\textcolor{ForestGreen}{Brushing Teeth} \\ $Energy=13.4$}
%  	\end{minipage}
%  	\hfill
%  	\begin{minipage}{0.3\columnwidth}
%  		\centerline{\includegraphics[width=\linewidth]{visualize/BrushingTeeth/BlowDryHair-11'3}}
%  		\subcaption{\textcolor{BrickRed}{Blow Dry Hair} \\$Energy=11.3$}
%  	\end{minipage}
%  \end{minipage}
%  \hfill

There are several findings from the results.
First, fine-tuning the CNN on the training videos significantly improves the performance,
and the improvement increases as more videos are used.
Second, the proposed method (EnergyNet) outperforms the baselines using different number of training videos.
The improvement is largest ($+3.1\%$) with 20\% of videos used.
When using all training videos, the improvement is $+0.7\%$.
This suggests that the proposed method is most effective when the amount of training data in the target domain is limited,
which is the general scenario for domain adaptation problems.
Furthermore, the proposed EnergyNet can achieve better performance using fewer videos compared with the baseline CNN method using more videos. 
(EnergyNet-10\% outperforms CNN-20\%, EnergyNet-50\% outperforms CNN-100\%)

\noindent\textbf{Comparison with state-of-the-art}.
In \tab~\ref{tbl:action_state} we compare our method with state-of-the-art results.
We directly quote the results from published papers.
Among those approaches, two-stream network~\cite{Simonyan_NIPS_2014}, IDT+FV~\cite{Wang_ICCV_2013}, C3D~\cite{Tran_ICCV_2015}, TDDs~\cite{Wang_CVPR_2015} and Ma~\etal spatial+IDT-FV~\cite{Ma_PR_2017} incorporate motion features from videos.
Other approaches only utilize appearance features from static frames.

Ma~\etal spatial~\cite{Ma_PR_2017} and Webly-supervised~\cite{Gan_ECCV_2016} are the two methods that most relate to ours.
Ma~\etal spatial~\cite{Ma_PR_2017} use images from the same BU101 dataset and all videos from UCF101 to train a shared CNN.
Webly-supervised~\cite{Gan_ECCV_2016} use images and videos from the Web to train a LSTM classifier.

With only 20\% of the training videos used, 
our method outperforms previous methods that only use spatial features, 
which again shows the efficacy of the proposed method when training data in the video domain is limited.
Using all training videos, our method achieves comparable performance with state-of-the-art methods that utilize both spatial and motion features.

\subsection{Visualization of Examples}
To provide more intuitions of how transferring attention contribute to video recognition, 
we show in \fig\ref{fig:visualization} some example frames from UCF101 where the baseline CNN method predicts wrongly and the proposed EnergyNet predicts correctly.
We show the spatial attention maps generated by the fine-tuned CNN for both the ground truth concept and the concept wrongly predicted by the baseline. 
For each example, the EnergyNet correctly assigns a higher energy to the spatial attention map corresponding to the ground truth concept.
 
\section{Conclusion}
\label{sec:conclusion}
In this work, we propose a new attention-based method to adapt a Web image trained CNN to video recognition.
The proposed method utilizes class-discriminative spatial attention map, which is a low-dimensional feature space that incorporates the convolutional features with class information.
We study unsupervised and supervised domain adaptation, 
and construct a Siamese EnergyNet that jointly optimizes two loss functions to learn class-specific energy functions for the attention maps.
We conduct experiments on human interaction recognition and action recognition,
and show the efficacy of the proposed method to adapt the domain shift problem,
especially when the amount of training data in the target domain is limited.

Since the proposed method focuses on static visual knowledge transfer from Web images to videos, 
we do not consider motion features.
For future work, 
we intend to incorporate motion features into our framework,
so that the performance can be further improved.
In addition, we believe that attention-based cross-domain knowledge transfer has other potentials beyond video recognition.
We aim to explore using Web images for action localization in videos, both spatially and temporally.

\section*{Acknowledgment}
\label{sec:acknowledgement}

This research is supported by the National Research Foundation, 
Prime Minister's Office, 
Singapore under its International Research Centre in Singapore Funding Initiative.
 
% \small

\balance
\bibliographystyle{ACM-Reference-Format}
\bibliography{reference}

\end{document}